\documentclass[runningheads,a4paper]{llncs}
\usepackage[T1]{fontenc}
\usepackage{amsmath,amssymb}
\usepackage{newtxtext}
\usepackage[varvw]{newtxmath}
\usepackage{graphicx}
\usepackage{array}
\usepackage{booktabs}
\usepackage{multirow}
\usepackage{adjustbox}
\usepackage[table]{xcolor}
\usepackage{microtype}
\usepackage{float}
\usepackage{subcaption}
\usepackage{tcolorbox}

\usepackage{wrapfig}

\usepackage[colorlinks=true,linkcolor=blue,citecolor=blue,urlcolor=blue,filecolor=blue]{hyperref}
\hypersetup{pdftitle={GeoMetric: Injecting Metric Geographic Structure into Worldwide Image Geo-Localization}}

\begin{document}
\title{GeoMetric: Injecting Metric Geographic Structure into Worldwide Image Geo-Localization}
\titlerunning{GeoMetric}
\author{Junchao Cui \and Wenqi Shi \and Xiangyang Luo \and
	Biru Zhu \and Xuanzi Ma \and Nan Wu }
\authorrunning{J. Cui et al.}
\institute{Information Engineering University}
\maketitle
\begingroup
\let\thefootnote\relax
\footnotetext{Preprint.}
\endgroup

\begin{abstract}
	Worldwide image geo-localization estimates where a photograph was taken on Earth. Existing methods infer location from visual appearance alone, an unreliable cue for geography: near-identical scenes occur on different continents. We present GeoMetric, a three-phase retrieval framework that encodes geographic distance directly in the representation. \textit{Phase~I} trains image, text, and location encoders into a unit hypersphere. The location encoder combines ScaleFusion and GeoAnchorAttention. ScaleFusion projects a coordinate onto an equal-area chart and encodes it with three Fourier bands, fusing them by attention into one multi-scale embedding. GeoAnchorAttention adds a global prior from fixed spherical anchors without mixing samples, so each coordinate embedding is batch-independent. A trimodal contrastive objective aligns images, reverse-geocoded texts, and coordinates, masking semantic and geographic false negatives. \textit{Phase~II} retrieves candidates, each carrying a GPS coordinate and granularity-aware geographic text. \textit{Phase~III} lets a frozen open-source 8B-parameter large multimodal model (LMM) choose the coordinate from those candidates. On four benchmarks, GeoMetric is best in 15 of 20 threshold cells, and ablations show that the components act at different scales: the text encoder carries coarse geographic context, while the location encoder resolves fine-grained distinctions.
\end{abstract}

\vspace{-10pt}
\section{Introduction}
\label{sec:intro}

Worldwide image geo-localization infers where on Earth a photograph was taken,
without metadata on the query and without a known search region. Most web images
carry no geotag, so the task underpins applications from source tracing to
disaster response and location-aware search~\cite{Choi22}. A recent
survey~\cite{Cui27} groups worldwide methods into four paradigms: geo-cell
classification~\cite{GeoDecoder23}, gallery retrieval~\cite{IM2GPS08,GeoCLIP23},
direct prediction by a large multimodal model
(LMM)~\cite{GeoReasoner,GRE}, and retrieval-augmented generation (RAG), which
prompts an LMM with retrieved coordinates~\cite{Img2loc24,G325,SpotAgent}.

\begin{figure}[t]
	\centering
	\includegraphics[width=0.6\linewidth]{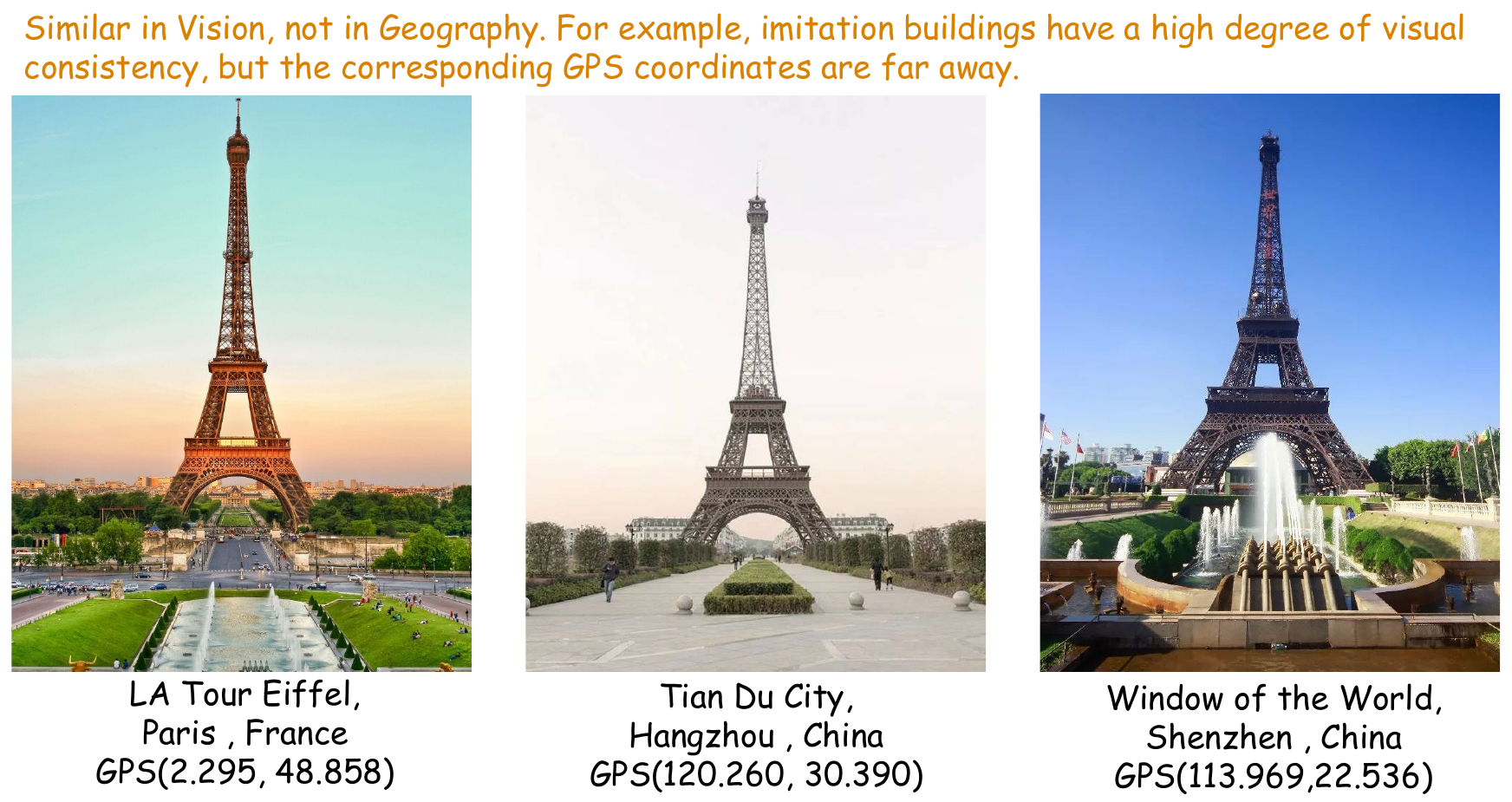}
	\vspace{-2pt}
	\caption{Visual ambiguity: near-identical buildings whose coordinates are far
		apart.}
	\label{fig:ambig}
	\vspace{-2pt}
\end{figure}

These paradigms differ in how a prediction is produced, yet all rest on one
premise: appearance determines location. The premise fails where the task is
hardest. Fig.~\ref{fig:ambig} shows three towers with a near-identical
silhouette, standing in Paris, Hangzhou and Shenzhen, eleven thousand kilometers
apart; a representation learned from appearance places them close together, so
the nearest neighbor of a query is as likely to be a replica as the original.
We call this \emph{visual ambiguity}; it dominates street-level accuracy, where
the prediction must separate look-alike candidates.

Prior efforts inject geography outside the representation. GeoCLIP~\cite{GeoCLIP23} aligns images and GPS coordinates through contrastive learning; G3~\cite{G325} converts coordinates into reverse-geocoded text; Img2Loc~\cite{Img2loc24} puts retrieved coordinates in the prompt; PIGEON~\cite{Haas24} smooths labels by Haversine distance; DualGeo~\cite{DualGeo} incorporates semantic segmentation into the gallery view. In each case distance modifies a label, a prompt or a reranking score, while the coordinate embedding remains a free vector: nothing ties it to the geometry of the sphere, and two coordinates a few kilometers apart can sit farther apart than two on different continents. We therefore change the encoder itself, not the supervision around it.

GeoMetric answers a query in three phases (Fig.~\ref{fig:overview}). \textit{Phase~I}
trains three encoders into a unit hypersphere. The location encoder takes the
coordinate alone. \textbf{ScaleFusion} projects the coordinate onto the Equal Earth
chart, a bounded equal-area map on which a fixed bandwidth covers the same
ground area everywhere. Three frozen Fourier banks of increasing bandwidth lift
the chart point, and attention fuses the three band tokens into one embedding that
spans street to continental scale. The resulting embedding is local: it captures how far apart two coordinates are, not where they lie on the globe. \textbf{GeoAnchorAttention} then
supplies that global frame by distance-decayed attention over a fixed set of
spherical anchors. Because it never attends across samples, a coordinate maps
to the same embedding whatever the batch, so the gallery is indexed once and
served unchanged. A trimodal contrastive objective aligns images,
reverse-geocoded texts and coordinates in one metric, masking false negatives
from identical place names and city-scale coordinate pairs, so the image encoder
does not separate what the text encoder cannot name. \textit{Phase~II} retrieves a
shortlist of candidates, each paired with its GPS coordinate and
granularity-aware geographic text. \textit{Phase~III} lets a frozen open-source 8B LMM
choose the coordinate from those candidates. Our main contributions are:
\vspace{-2pt}
\begin{itemize}
	\item[$\bullet$] We design a metric location encoder: ScaleFusion fuses three spectral bands on an equal-area chart, and GeoAnchorAttention adds a global prior while keeping every coordinate embedding a per-sample function.
	\item[$\bullet$] We align images, texts and GPS coordinates with a trimodal contrastive objective whose masks follow the label tolerance, dropping duplicate captions and city-scale coordinate pairs so that one space serves image, text and GPS queries.
	\item[$\bullet$] We introduce training-free candidate-reasoning with granularity-aware text, and an open 8B model is best in 15 of 20 threshold cells on four benchmarks.
\end{itemize}

\section{Related Work}
\label{sec:related}

\subsection{Geographic Classification}
\label{sec:rw-cls}
Classification methods discretize the Earth and cast localization as multiclass
classification. PlaNet~\cite{PlaNet16} partitions the globe with the S2 grid;
CPlaNet~\cite{CPlaNet18} refines the partition by combinations of coarse cells;
ISNs~\cite{ISNs18} add a hierarchy with scene classification;
TransLocator~\cite{TransLocator22} fuses semantic segmentation maps; and
GeoDecoder~\cite{GeoDecoder23} attends over geographic hierarchies and scene
labels. The limitation is quantization: a correct cell still incurs an error of
its radius, and coarse cells cannot express street-level precision. Their
strength is complementary: a fixed partition is a strong prior at continental
scale, which is why they stay competitive at 750 and 2{,}500\,km.

\subsection{Image Retrieval}
\label{sec:rw-ret}
Retrieval methods match the query against a geotagged gallery. IM2GPS~\cite{IM2GPS08}
pioneered the paradigm; Vo et al.~\cite{Vo17} combine it with PlaNet through grid
partitioning and kernel density estimation; GeoCLIP~\cite{GeoCLIP23} aligns
images and coordinates with CLIP-style contrastive learning and random Fourier
features; PIGEON~\cite{Haas24} combines semantic partitioning with
clustering-balanced pre-training; HierLoc~\cite{HierLoc26} learns hyperbolic
embeddings of the geographic hierarchy; and DualGeo~\cite{DualGeo} retrieves with
two complementary views of the gallery. All of them learn one embedding per
coordinate from appearance-driven supervision, so a distant look-alike remains
the nearest neighbor. Serving such a gallery is a database problem in its own right, and it is studied in this community: distance-aware space partitions~\cite{DASFAA25ANN}, learned
spatial indexes~\cite{DASFAA25LASE}, spatio-temporal phrase
indexes~\cite{GeoPhraseTree} and streaming nearest-neighbor
processing~\cite{DASFAA25GNN}. GeoMetric is complementary to that line: it does
not change how the index searches, it changes what the index is asked to search,
namely a coordinate embedding that is a multi-scale, anchor-contextualized
function of the sphere rather than of the training batch.

\subsection{LMM-based and Retrieval-Augmented Localization} 
\label{sec:rw-lmm}
The remaining two paradigms put an LMM at the center and differ in what grounds it. Direct prediction prompts the model to reason from the image alone~\cite{GeoReasoner,GRE}, whether as a multi-agent system that infers the location by invoking external tools~\cite{SpotAgent} or as a sequential Bayesian posterior over candidates~\cite{GeoBayes}: it exploits world knowledge such as architecture, vegetation and traffic signs, but it is ungrounded and hallucinates for rare regions. Retrieval-augmented methods instead append retrieved evidence to the prompt, as the coordinate of the top match~\cite{Img2loc24}, as aligned image-GPS context~\cite{G325}, or as hierarchical geographic embeddings~\cite{GeoSURGE}; a related line trains a ranker over retrieved candidates~\cite{GeoRanker25}. The weakness is where geography is applied: distance enters at the prompt, at the posterior or at the reranking score, while the retrieval embedding stays appearance-driven, so the candidate pool itself is chosen by visual similarity. Multimodal fusion is studied in adjacent tasks as well~\cite{CAMeRA}, but geo-localization adds a constraint: the fused representation must respect a known metric on the output space. GeoMetric closes the gap from the other side: the candidates come from a metric representation, and the LMM stage only has to choose among them, which is why it can stay training-free.

\section{Methodology}
\label{sec:method}

\begin{figure}[t]
	\centering
	\includegraphics[width=\linewidth]{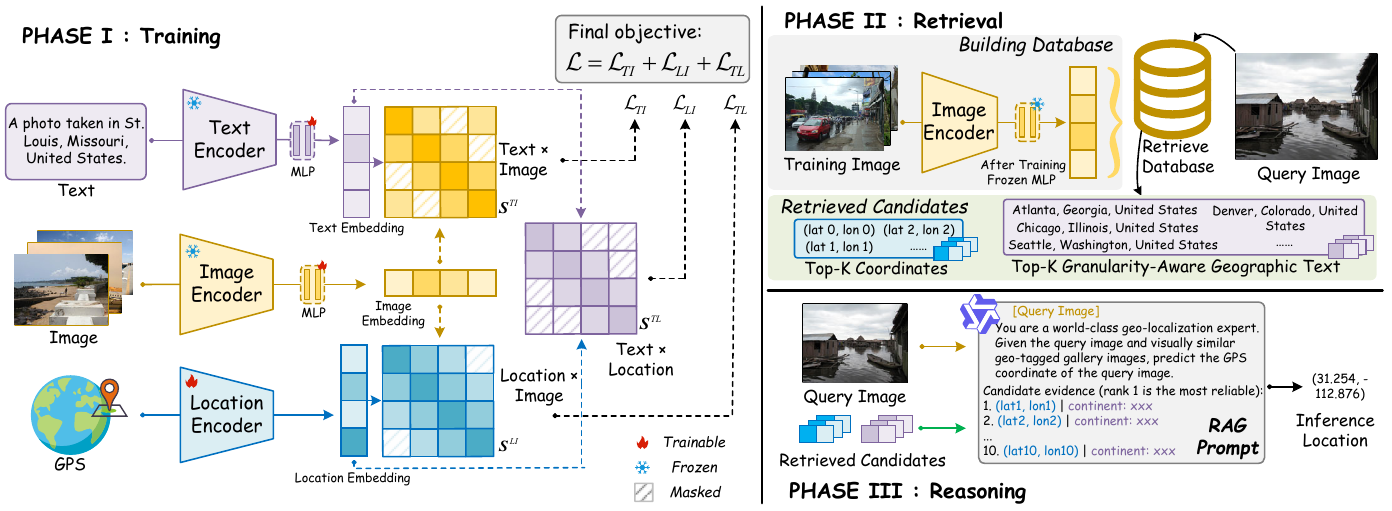}
	\caption{\textbf{Overview of GeoMetric.} \emph{Phase I} (training) aligns image, geographic-text and location embeddings in one unit hypersphere with three pairwise contrastive losses; hatched cells are similarities removed by the false-negative masks. The location encoder combines \emph{ScaleFusion}, which fuses random-Fourier-feature capsules of three bandwidths by scale attention, and \emph{GeoAnchorAttention}, which injects a global prior from a fixed anchor set. \emph{Phase II} (retrieval) builds the database once with the trained image encoder and returns the $k$ candidates of a query. \emph{Phase III} (reasoning) lets a frozen LMM produce the coordinate from those candidates and their granularity-aware geographic text, with no additional training and no ground-truth coordinate in the prompt.}
	\label{fig:overview}
\end{figure}

\subsection{Overview and Design Rationale}
\label{sec:overview}

We treat worldwide image geo-localization as retrieval over a geotagged gallery
of $N\!\approx\!4.1$M user-contributed photographs from social-media platforms
(MP16-Pro~\cite{G325}); each item comprises an image, its reverse-geocoded text
and a coordinate $\mathbf{y}$. A prediction is correct at threshold $\delta$ when its
great-circle error is below $\delta$, with
$\delta\in\{1,25,200,750,2500\}$\,km, i.e.\ from street to continent.

GeoMetric works in three phases (Fig.~\ref{fig:overview}). \emph{Phase I} learns
the image, text and location encoders, which map images, captions and
coordinates into one unit hypersphere with three pairwise contrastive losses. \emph{Phase II} encodes the
gallery once with the trained image encoder and returns the $k$ nearest
neighbors of a query; \emph{Phase III} renders those candidates for a frozen
LMM, which returns the coordinate. Only Phase I is trained: a query costs one
image-encoder pass, one inner-product search and, optionally, one LMM call.

Three requirements follow. \textbf{(R1) Scale.} Visual evidence localizes at
street, city or continental granularity, and no single bandwidth serves all
three, so the coordinate code must be multi-scale, bounded and pole-free.
\textbf{(R2) Consistency.} The same coordinate must map to the same embedding
whatever else is in the batch, because the database is built once and served to
every query; attention over batch-mates would make the code a function of batch
composition and undefined for a single query. \textbf{(R3) One metric under
noisy labels.} The three modalities must share one space, or their similarities
are incomparable and one index cannot serve image$\rightarrow$GPS,
text$\rightarrow$GPS and GPS$\rightarrow$image queries; and geotags contain
false negatives that a naive contrastive denominator actively repels.
Sects.~\ref{sec:scalefusion} and~\ref{sec:geoanchor} build the location encoder
for (R1) and (R2), Sect.~\ref{sec:trimodal} the shared space for (R3), and
Sect.~\ref{sec:inference} uses it without further training.

\begin{figure}[t]
	\centering
	\includegraphics[width=\linewidth]{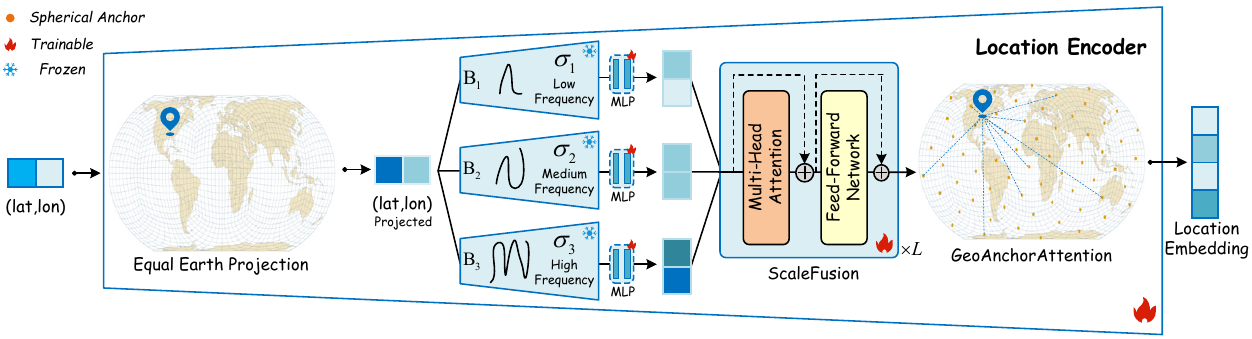}
	\caption{\textbf{The location encoder.} A coordinate is projected onto the Equal Earth chart and lifted by three frozen random-Fourier-feature banks of low, medium and high frequency; one trainable capsule MLP per band yields a token, and \emph{ScaleFusion} fuses the three tokens with two stacked attention and feed-forward blocks. \emph{GeoAnchorAttention} updates the code by distance-decayed attention over a fixed set of spherical anchors, and a final projection gives the location embedding.}
	\label{fig:location}
\end{figure}

\subsection{ScaleFusion: Multi-Scale Encoding on an Equal-Area Chart}
\label{sec:scalefusion}

The location encoder (Fig.~\ref{fig:location}) lifts the coordinate with Fourier
features, which are defined on the plane and therefore need a planar chart,
and the chart fixes what a bandwidth means: it must be bounded and pole-free,
and a fixed bandwidth must cover roughly the same ground area everywhere,
because the three bands denote three ground scales. Mercator, the chart of
G3~\cite{G325}, satisfies neither requirement: $y_e$ diverges at the
poles, and its area scale grows as $\sec^2\phi$, so a band that spans a street
at the equator spans a province at $60^\circ$. We use the Equal Earth projection~\cite{Savric19}, a bounded,
equal-area pseudocylindrical chart,
\begin{equation}
\begin{aligned}
	\theta&=\arcsin\!\left(\tfrac{\sqrt{3}}{2}\sin\phi\right),\qquad
	y_e=\theta\left(A_1+A_2\theta^2+\theta^6\left(A_3+A_4\theta^2\right)\right),\\
	x_e&=\frac{2\sqrt{3}}{3}\,\lambda\,
	\frac{\cos\theta}{A_1+3A_2\theta^2+\theta^6\left(7A_3+9A_4\theta^2\right)},
\end{aligned}
\label{eq:eep}
\end{equation}
with $A_1{=}1.340264$, $A_2{=}-0.081106$, $A_3{=}0.000893$ and
$A_4{=}0.003796$; dividing both axes by their analytic extrema gives
$\mathbf{p}\in[-1,1]^2$. Since the area scale is constant, a chart kernel of
fixed bandwidth covers a spherical neighborhood of roughly constant area at
every latitude, so a band denotes a ground scale rather than a latitude. The
chart only parameterizes the features: every distance below is a great-circle
distance,
\begin{equation}
	d(\mathbf{y},\mathbf{y}')=2R\arcsin\sqrt{
	\sin^2\tfrac{\Delta\phi}{2}
	+\cos\phi\cos\phi'\sin^2\tfrac{\Delta\lambda}{2}},
	\label{eq:haversine}
\end{equation}
with $R{=}6371.0088$\,km, so no planar distortion enters the modeled geometry.
\vspace{-3pt}
\paragraph{One capsule per band.} Scale $s$ owns a frequency matrix
$\mathbf{B}_s\in\mathbb{R}^{m\times2}$, drawn once from a zero-mean Gaussian
with variance $\sigma_s^2$ and then frozen, and a capsule that lifts the chart
coordinate,
\begin{equation}
	\gamma_s(\mathbf{p})
	=\left[\cos\left(2\pi\mathbf{p}\mathbf{B}_s^{\top}\right),\;
	\sin\left(2\pi\mathbf{p}\mathbf{B}_s^{\top}\right)\right],
	\qquad
	\mathbf{h}_s=\mathrm{LN}\!\left(\mathrm{MLP}_s(\gamma_s(\mathbf{p}))\right)\in\mathbb{R}^{d},
	\label{eq:capsule}
\end{equation}
with $m{=}256$, $d{=}512$, $S{=}3$ bands $\sigma_s\in\{1,16,256\}$, i.e.\
$2^0,2^4,2^8$, and an MLP of width $1024$. Because the basis is frozen, the code
is a well-defined function of position while the MLPs learn which part of the
spectrum to use. By the random-feature argument~\cite{Rahimi07},
$\mathbb{E}\big[\tfrac{1}{m}\langle\gamma_s(\mathbf{p}),
\gamma_s(\mathbf{p}')\rangle\big]
=\exp(-2\pi^2\sigma_s^2\|\mathbf{p}-\mathbf{p}'\|_2^2)
=:k_{\sigma_s}(\mathbf{p}-\mathbf{p}')$, so the concatenated map induces
$\sum_s k_{\sigma_s}$: a positive sum of characteristic kernels, hence itself
characteristic, which exposes street, regional and continental scales at once.
\vspace{-3pt}
\paragraph{Scale fusion.} The capsule tokens $\mathbf{h}_s$ form a short sequence, and a scale
embedding identifies the band of a token and breaks the permutation symmetry. A
two-layer pre-norm Transformer~\cite{Vaswani17} with $8$ heads, width $1024$ and
dropout $0.1$ fuses the sequence; averaging its output tokens gives
\begin{equation}
	\mathbf{z}=\mathrm{LN}\!\left(\tfrac{1}{S}\textstyle\sum_{s=1}^{S}
	\mathbf{H}^{(2)}_{s}\right),
	\label{eq:fusion}
\end{equation}
where $\mathbf{H}^{(2)}_{s}$ is the output token of band $s$. Without attention,
$\mathbf{z}$ is a fixed linear combination of the band features, and the induced
kernel contains the multi-bandwidth Gaussian blocks and the cross-band
correlations; attention makes the mixing weights input-dependent, so a street
scene can lean on the fine band and a desert on the coarse one, at the cost of
six cross-band attention weights. The induced kernel is then input-dependent
rather than a fixed sum of per-band kernels.

\subsection{GeoAnchorAttention: A Global Prior Anchored on the Sphere}
\label{sec:geoanchor}

ScaleFusion describes a coordinate by its position in a chart
(Fig.~\ref{fig:location}), and its code is a local kernel feature: the inner
product of two codes decays with distance and is close to zero at continental
separation, so the representation says how far apart two places are, not which
part of the world they occupy. Recovering the latter from the spectrum alone
leaves the MLP without a shared reference frame where data are sparse, so we add
a global prior: a partition of the sphere learned once and shared by all
coordinates.
\vspace{-5pt}
\paragraph{Why not attend over the batch.} The obvious context is the rest of
the batch, and we reject it for (R2): the embedding of a coordinate would depend
on its batch-mates, change at every step and be undefined for a single query.
The database is built once and served unchanged, so the location encoder must be
a per-sample function.
\vspace{-5pt}
\paragraph{Anchors and distance-decayed attention.} We place $K{=}64$ anchors by
a Fibonacci lattice~\cite{Gonzalez10}, which covers the sphere quasi-uniformly
with covering radius $O(K^{-1/2})$, and define
\begin{equation}
	\mathbf{c}(\mathbf{y})=\sum_{k=1}^{K}
	\frac{e^{-d_k(\mathbf{y})/\tau}}
	{\sum_{j=1}^{K}e^{-d_j(\mathbf{y})/\tau}}\,\mathbf{e}_k,
	\label{eq:context}
\end{equation}
with $d_k$ from Eq.~\eqref{eq:haversine}, learnable anchor embeddings
$\mathbf{e}_k\in\mathbb{R}^{d}$ and a positive learned temperature $\tau$
initialized at $2000$\,km. The code is updated
by a gated residual and projected to the shared space,
\begin{equation}
	\mathbf{z}'=\mathrm{LN}\!\left(\mathbf{z}+\tanh(\beta)\,\mathbf{c}(\mathbf{y})\right),
	\qquad
	\mathbf{w}=\frac{g_L(\mathbf{z}')}{\|g_L(\mathbf{z}')\|_2},
	\label{eq:geoanchor}
\end{equation}
where $\mathbf{w}$ is the location embedding, $\beta$ is a gate initialized to
zero and $g_L$ is a two-layer MLP. The context is a kernel-smoothed lookup of a
learnable field, so nearby coordinates receive similar mixtures: the embedding
inherits a smooth spatial prior and a shared global frame. The gate starts at
$\tanh(0){=}0$, so the module is the identity at initialization and the prior
enters gradually instead of corrupting the code early in training.
\vspace{-5pt}
\paragraph{What the module provides.} The attention never crosses samples, so a
coordinate receives the same embedding in any batch that contains it, and its
embedding is well defined for a batch of one;
since each anchor distance is 1-Lipschitz,
$\|\mathbf{c}(\mathbf{y})-\mathbf{c}(\mathbf{y}')\|_2\le
(K\max_k\|\mathbf{e}_k\|_2/\tau)\,d(\mathbf{y},\mathbf{y}')$, so neighbors
cannot receive arbitrarily different contexts. The temperature interpolates
between a hard regional lookup ($\tau\to0$) and a constant bias
($\tau\to\infty$). 

\subsection{Trimodal Contrastive Alignment}
\label{sec:trimodal}
\paragraph{Encoders.} CLIP ViT-L/14-336~\cite{Clip21} stays frozen: for an image
and its caption it yields
$\mathbf{e}^{I},\mathbf{e}^{T}\in\mathbb{R}^{768}$, and two-layer MLPs
$g_I,g_T$ (GELU, hidden size $512$) with $\ell_2$ normalization give
$\mathbf{v}=g_I(\mathbf{e}^{I})/\|g_I(\mathbf{e}^{I})\|_2$ and
$\mathbf{u}=g_T(\mathbf{e}^{T})/\|g_T(\mathbf{e}^{T})\|_2$. The location encoder
is trained from scratch in the same $d$-dimensional space. Freezing preserves
CLIP's semantic metric against noisy geotags and lets the gallery features be
precomputed once; $12.4$M parameters are trainable, $11.1$M of them in the
location encoder.
\vspace{-5pt}
\paragraph{Captions.} Following G3~\cite{G325}, we obtain the text of an item by
reverse-geocoding its coordinate: $(38.63,-90.20)$ becomes ``\emph{A photo taken
in St. Louis, Missouri, United States}'', which names the city, its state and
the country of the point. This is a coarse place name rather than a description
of the scene, and it gives the text encoder a geographic signal that the location
encoder lacks.
\vspace{-5pt}
\paragraph{Objective.} With a learnable scale $\kappa$ clamped at $100$ and
initialized at $1/0.07$, the three similarity matrices and the symmetric
InfoNCE loss are
\begin{align}
	\mathbf{S}^{TI}_{ij}&=\kappa\,\mathbf{u}_i^{\top}\mathbf{v}_j,\quad
	\mathbf{S}^{LI}_{ij}=\kappa\,\mathbf{w}_i^{\top}\mathbf{v}_j,\quad
	\mathbf{S}^{TL}_{ij}=\kappa\,\mathbf{u}_i^{\top}\mathbf{w}_j,\label{eq:similarity}\\
	\mathcal{L}(\mathbf{S},M)&=-\frac{1}{2B}\sum_{i=1}^{B}\left[
	\log\frac{e^{\mathbf{S}_{ii}}}{\sum_{j:M_{ij}=0}e^{\mathbf{S}_{ij}}}
	+
	\log\frac{e^{\mathbf{S}_{ii}}}{\sum_{j:M_{ij}=0}e^{\mathbf{S}_{ji}}}
	\right],\label{eq:infonce}
\end{align}
where $M$ is the mask of the term and $M_{ij}=1$ marks a false negative. The text
mask sets $M^{T}_{ij}=1$ when the captions of $i$ and $j$ share a CRC32 hash, so
it groups images that share one place name; the geographic mask sets
$M^{G}_{ij}=1$ for pairs whose coordinates are closer than $\delta{=}25$\,km, the
city-level threshold of the evaluation protocol. The text-image loss is
$\mathcal{L}_{TI}=\mathcal{L}(\mathbf{S}^{TI},M^{T})$, the location-image loss is
$\mathcal{L}_{LI}=\mathcal{L}(\mathbf{S}^{LI},M^{G})$, and the text-location loss
is $\mathcal{L}_{TL}=\mathcal{L}(\mathbf{S}^{TL},M^{T}\cup M^{G})$, with all masks
symmetric and without the diagonal; the final objective is
$\mathcal{L}=\mathcal{L}_{TI}+\mathcal{L}_{LI}+\mathcal{L}_{TL}$.
\vspace{-3pt}
\paragraph{Why the masks.} They encode the tolerance of the label: a caption is
a place name, not a point, so every image that shares it is textually
indistinguishable and the text encoder cannot separate two of them; two captures
within $\delta$ are, up to capture geometry and geotag noise, the same place.
Masking such pairs keeps the image encoder from separating what the text encoder
cannot name, which is the debiasing of contrastive learning~\cite{Chuang20} at
the semantic and the geometric level.
\emph{One space.} The three terms share their embeddings and hence one metric; each
lower-bounds the mutual information of its pair, the shared metric
rules out three incomparable subspaces, and one index answers all three query
directions.

\subsection{Contrastive-Candidate Inference}
\label{sec:inference}

\paragraph{Building the database.} The trained image encoder is applied once to
the gallery, and its $\ell_2$-normalized outputs go into an exact FAISS
inner-product index~\cite{Johnson19}. A query returns the $k$ gallery items with
the largest inner products, together with their reverse-geocoded records. The
index stores $N\times d$ float32 vectors and adds neither
parameters nor training; the candidates come from the contrastively learned
space and are presented as a ranked set among which the model must discriminate.
\vspace{-6pt}
\paragraph{Granularity-aware geographic text.} Each candidate is rendered as its
coordinate followed by up to three fields: city, country and continent. The
fields are selected by a granularity level
$g\in\{\text{none},\text{continent},\text{country},\text{city},\text{all}\}$:
$g{=}\text{none}$ keeps the coordinate only, and $g{=}\text{all}$ adds the three
names. A field that reverse geocoding does not return is omitted. Coarse labels
resist geocoding noise, and the city label is discriminative; once it is
missing, the coordinate keeps the candidates grounded even at
$g{=}\text{none}$.
\vspace{-6pt}
\paragraph{Prompt and decoding.} The prompt ranks the candidates, states that
rank~1 is the most reliable, asks the model to check architecture, vegetation,
road signs, language, terrain and lighting, and to return one coordinate
as JSON. The generator is a frozen Qwen3-VL-8B-Instruct~\cite{Qwen3VL} with
greedy decoding and at most $64$ new tokens. An answer is accepted only if it
parses to a finite in-range coordinate. The ground truth never appears in the
prompt. For illustration, the prompt below uses $g{=}\text{city}$ and $k{=}10$:

\begingroup
\setlength{\fboxsep}{5pt}%
\setlength{\fboxrule}{0.5pt}%
\noindent
\fcolorbox{black!35}{gray!6}{%
	\begin{minipage}{\dimexpr\linewidth-2\fboxsep-2\fboxrule\relax}
		\vspace{1pt}%
		\noindent
		\begin{minipage}[t]{0.48\linewidth}
			\scriptsize\ttfamily\raggedright
			You are a world-class geo-localization expert. Given the query image and
			visually similar geo-tagged gallery images, predict the GPS coordinate of the
			query image.\\[2pt]
			Candidate evidence (rank 1 is the most reliable):\\
			1. \textcolor{blue!75!black}{(lat1, lon1)} | \textcolor{purple!80!black}{City}: xxx\\
			2. \textcolor{blue!75!black}{(lat2, lon2)} | \textcolor{purple!80!black}{City}: xxx\\
			\dots\\
			10. \textcolor{blue!75!black}{(lat10, lon10)} | \textcolor{purple!80!black}{City}: xxx\\
			Candidate city text is noisy; use it as a coarse prior.
		\end{minipage}\hfill
		\begin{minipage}[t]{0.48\linewidth}
			\scriptsize\ttfamily\raggedright
			Follow these steps:\\
			1. Observe visual clues: architecture, vegetation, road signs, language,
			terrain, lighting.\\
			2. Use the candidate cluster: the true location is likely near top-ranked
			candidates.\\
			3. If candidates spread across continents, select the cluster supported by
			visual cues.\\
			4. If uncertain, lean toward the center of the most likely continent.\\[2pt]
			Output ONLY a JSON object: \{"latitude": float, "longitude": float\}
		\end{minipage}
		\vspace{1pt}%
\end{minipage}}%
\endgroup
\vspace{-5pt}
\paragraph{Why candidate reasoning helps.} The retriever induces a prior over
the $k$ candidates from their retrieval similarities, and the LMM induces a
conditional distribution over coordinates given the image, a candidate and its
geographic text; the pipeline is their mixture, whose mode is the answer, with
the prompt implementing the weights through the ranking. Hence $k$ sets the
recall of the support and $g$ the informativeness of the evidence, and the stage
is training-free and model-agnostic: it changes what the LMM sees, not what the
LMM is.
\vspace{-5pt}

\section{Experiments}
\label{sec:exp}

\newcommand{\ck}{\textcolor{green!55!black}{$\checkmark$}}
\newcommand{\cx}{\textcolor{red}{$\times$}}

We organize the evaluation around five questions: (i) how GeoMetric compares with state-of-the-art approaches at every geographic level across all four paradigms (Sect.~\ref{sec:cmp}); (ii) how much gain resides in the retrieval representation before LMM involvement (Sect.~\ref{sec:retr}); (iii) the contribution of each individual component (Sect.~\ref{sec:abl}); (iv) inference sensitivity to candidate budget, geographic text and LMM backbone (Sect.~\ref{sec:inf}); and (v) where the geographic structure of the shared space originates (Sect.~\ref{sec:vis}).

\paragraph{Datasets.} We train and index on MP16-Pro~\cite{G325}: 4.1M
photographs from social-media platforms, each with the reverse-geocoded text of
Sect.~\ref{sec:trimodal}. Evaluation uses four benchmarks: IM2GPS3k~\cite{Vo17}, YFCC4k~\cite{Vo17} (4,368 images after de-duplication), the
hand-curated IM2GPS~\cite{IM2GPS08} and YFCC26k~\cite{ISNs18}.

\paragraph{Metrics.} We report the fraction of predictions whose geodesic error falls below $\delta\in\{1,25,200,750,2500\}$\,km, which correspond to street, city, regional, country, and continent granularities, alongside the median error in kilometers.

\paragraph{Implementation.} The image and text encoders use CLIP
ViT-L/14-336~\cite{Clip21} with the backbone frozen; the location encoder uses
$d{=}512$, $m{=}256$, $S{=}3$ bands, and $K{=}64$ anchors, giving $12.4$M trainable parameters. We train for 10 epochs with AdamW
(learning rate $5\times10^{-5}$, weight decay $10^{-6}$), cosine decay with a
$5\%$ warm-up and bf16, on two A6000 GPUs with 1024 images each (2048 per step).
Retrieval is an exact FAISS inner-product search over the 4.1M gallery
embeddings. Unless stated otherwise, the LMM is
Qwen3-VL-8B-Instruct~\cite{Qwen3VL} with $k{=}10$ candidates, greedy decoding
and at most 64 new tokens.

\paragraph{Baselines.} We compare our method against a set of prior state-of-the-art approaches, including PlaNet~\cite{PlaNet16}, ISNs~\cite{ISNs18}, Semp~\cite{Semp22}, TransLocator~\cite{TransLocator22}, GeoDecoder~\cite{GeoDecoder23}, PIGEON~\cite{Haas24}, HierLoc~\cite{HierLoc26}, GeoCLIP~\cite{GeoCLIP23}, GeoReasoner~\cite{GeoReasoner}, G3~\cite{G325}, SpotAgent~\cite{SpotAgent}, GeoBayes~\cite{GeoBayes}, DualGeo~\cite{DualGeo}, and GeoSURGE~\cite{GeoSURGE}.

\begin{table}[t]
	\centering
	\caption{\textbf{Comparison on IM2GPS3k and YFCC4k.} Best in \textbf{bold}, second best \underline{underlined}. Higher is better. The baseline results are as reported by the original authors.}
	\label{tab:main1}
	\scriptsize
	\vspace{2pt}
	\renewcommand{\arraystretch}{1.1}
	\setlength{\tabcolsep}{4pt}
	\begin{tabular}{@{}lccccc|ccccc@{}}
		\toprule
		\multirow{2}{*}{\textbf{Method}} & \multicolumn{5}{c}{\textbf{IM2GPS3k}} & \multicolumn{5}{c}{\textbf{YFCC4k}} \\
		\cmidrule(lr){2-6}\cmidrule(lr){7-11}
		& 1\,km & 25\,km & 200\,km & 750\,km & 2500\,km & 1\,km & 25\,km & 200\,km & 750\,km & 2500\,km \\
		\midrule
		PlaNet~\cite{PlaNet16} & 8.5 & 24.8 & 34.3 & 48.4 & 64.6 & 5.6 & 14.3 & 22.2 & 36.4 & 55.8 \\
		ISNs~\cite{ISNs18} & 10.5 & 28.0 & 36.6 & 49.7 & 66.0 & 6.5 & 16.2 & 23.8 & 37.4 & 55.0 \\
		Semp~\cite{Semp22} & 12.5 & 31.4 & 42.7 & 57.3 & 72.0 & 9.4 & 20.3 & 30.6 & 44.8 & 61.2 \\
		TransLocator~\cite{TransLocator22} & 11.8 & 31.1 & 46.7 & 58.9 & 80.1 & 8.4 & 18.6 & 27.0 & 41.1 & 60.4 \\
		GeoDecoder~\cite{GeoDecoder23} & 12.8 & 33.5 & 45.9 & 61.0 & 76.1 & 10.3 & 24.4 & 33.9 & 50.0 & 68.7 \\
		PIGEON~\cite{Haas24} & 11.3 & 36.7 & 53.8 & 72.4 & 85.3 & 10.4 & 23.7 & 40.6 & 62.2 & 77.7 \\
		GeoReasoner~\cite{GeoReasoner} & 8.1 & 28.8 & 43.3 & 62.9 & 80.7 & 3.7 & 17.0 & 28.5 & 50.6 & 67.3 \\
		G3~\cite{G325} & 16.7 & 40.9 & 55.6 & 71.2 & 84.7 & 24.0 & 35.9 & \underline{47.0} & \underline{64.3} & \underline{78.2} \\
		SpotAgent~\cite{SpotAgent} & 14.1 & 40.4 & \textbf{57.8} & \underline{73.4} & 85.8 & 7.3 & 21.5 & 36.2 & 55.0 & 70.8 \\
		GeoBayes~\cite{GeoBayes} & 6.3 & 34.7 & 53.6 & \textbf{73.7} & \underline{85.9} & 4.9 & 16.1 & 30.9 & 55.8 & 75.4 \\
		DualGeo~\cite{DualGeo} & \textbf{17.3} & \underline{41.5} & 55.8 & 71.7 & 85.1 & \textbf{27.5} & \underline{36.5} & 45.0 & 61.6 & 75.9 \\
		\midrule
		GeoMetric (ours) & \underline{17.1} & \textbf{41.7} & \underline{56.2} & \textbf{73.7} & \textbf{86.4} & \underline{27.3} & \textbf{38.6} & \textbf{49.8} & \textbf{67.2} & \textbf{81.3} \\
		\bottomrule
	\end{tabular}
\end{table}

\subsection{Comparison with State-of-the-Art Methods}
\label{sec:cmp}

Tables~\ref{tab:main1} and~\ref{tab:main2} report the four benchmarks. GeoMetric
is first in 15 of the 20 threshold cells; in the remaining five it is second or third. DualGeo~\cite{DualGeo} accounts for the largest share of the remaining first places, with SpotAgent~\cite{SpotAgent} and GeoDecoder~\cite{GeoDecoder23} leading one cell each. GeoMetric uses an open 8B LMM, while two of the strongest baselines call closed-source models (G3~\cite{G325} uses GPT-4V and DualGeo~\cite{DualGeo} uses Qwen3-VL-Plus).

At 1\,km, GeoMetric is the strongest open-model method on all four benchmarks.
DualGeo is narrowly ahead on two of them, and GeoMetric is first on the other
two. On IM2GPS3k and YFCC4k, GeoMetric matches or exceeds DualGeo~\cite{DualGeo} from 25\,km
upward. It is also ahead of G3~\cite{G325} at every threshold that G3 reports,
and at 2{,}500\,km it leads on all four benchmarks; on YFCC26k it leads at every
threshold.

Of the five, two are at 1\,km, where DualGeo~\cite{DualGeo} is narrowly ahead; the rest are at city-to-region scales, where methods with explicit geographic priors (fixed
partitions, hierarchies, or agentic planning) remain competitive. These cells
qualify rather than reverse the overall picture: the representation is designed
for candidate-level discrimination, while explicit geographic priors remain
competitive for region-level decisions.

\begin{table}[t]
	\centering
	\caption{\textbf{Comparison on (a) IM2GPS and (b) YFCC26k.} Best in \textbf{bold}, second best \underline{underlined}. Higher is better. The baseline results are as reported by the original authors.}
	\label{tab:main2}
	\begin{subtable}[t]{0.48\linewidth}
		\centering
		\scriptsize
		\vspace{2pt}
		\renewcommand{\arraystretch}{1.05}
		\setlength{\tabcolsep}{1.8pt}
		\begin{tabular}{@{}lccccc@{}}
			\toprule
			\multicolumn{6}{c}{\textbf{IM2GPS}} \\
			\cmidrule(lr){1-6}
			\textbf{Method} & 1\,km & 25\,km & 200\,km & 750\,km & 2500\,km \\
			\midrule
			PlaNet~\cite{PlaNet16} & 8.4 & 24.5 & 37.6 & 53.6 & 71.3 \\
			ISNs~\cite{ISNs18} & 16.9 & 43.0 & 51.9 & 66.7 & 80.2 \\
			Semp~\cite{Semp22} & 16.9 & 42.6 & 56.1 & 69.6 & 84.8 \\
			TransLocator~\cite{TransLocator22} & 19.9 & 48.1 & 64.6 & 75.6 & 86.7 \\
			GeoDecoder~\cite{GeoDecoder23} & 22.1 & 50.2 & \textbf{69.0} & 80.0 & 89.1 \\
			PIGEON~\cite{Haas24} & 14.8 & 40.9 & 63.3 & 82.3 & 91.1 \\
			HierLoc~\cite{HierLoc26} & 10.5 & \underline{51.9} & \underline{67.5} & \underline{83.1} & \underline{92.4} \\
			DualGeo~\cite{DualGeo} & \underline{23.2} & \textbf{52.3} & 65.4 & 78.9 & 89.9 \\
			\midrule
			GeoMetric (ours) & \textbf{24.9} & 51.1 & 66.7 & \textbf{83.5} & \textbf{93.7} \\
			\bottomrule
		\end{tabular}
		\caption{IM2GPS.}
	\end{subtable}\hfill
	\begin{subtable}[t]{0.48\linewidth}
		\centering
		\scriptsize
		\vspace{2pt}
		\renewcommand{\arraystretch}{1.05}
		\setlength{\tabcolsep}{1.8pt}
		\begin{tabular}{@{}lccccc@{}}
			\toprule
			\multicolumn{6}{c}{\textbf{YFCC26k}} \\
			\cmidrule(lr){1-6}
			\textbf{Method} & 1\,km & 25\,km & 200\,km & 750\,km & 2500\,km \\
			\midrule
			PlaNet~\cite{PlaNet16} & 4.4 & 11.0 & 16.9 & 28.5 & 47.7 \\
			ISNs~\cite{ISNs18} & 5.3 & 12.3 & 19.0 & 31.9 & 50.7 \\
			Semp~\cite{Semp22} & 6.8 & 16.4 & 24.6 & 38.4 & 56.8 \\
			TransLocator~\cite{TransLocator22} & 7.2 & 17.8 & 28.0 & 41.3 & 60.6 \\
			GeoDecoder~\cite{GeoDecoder23} & 10.1 & 23.9 & 34.1 & 49.6 & 69.0 \\
			PIGEON~\cite{Haas24} & 10.1 & 24.6 & 41.3 & 62.6 & 78.7 \\
			GeoSURGE~\cite{GeoSURGE} & \underline{17.8} & \underline{31.5} & \underline{45.1} & \underline{64.3} & \underline{79.3} \\
			\midrule
			GeoMetric (ours) & \textbf{20.5} & \textbf{34.5} & \textbf{47.8} & \textbf{66.1} & \textbf{80.9} \\
			\bottomrule
		\end{tabular}
		\caption{YFCC26k.}
	\end{subtable}
\end{table}

\begin{table}[t]
	\centering
	\caption{\textbf{Retrieval-only comparison.} No LMM is used; every row is the
		top-1 retrieval of its own reproduced gallery. $^*$: rebuilt in our
		environment. Best in \textbf{bold}, second best \underline{underlined}.}
	\label{tab:ret}
	\scriptsize
	\vspace{2pt}
	\renewcommand{\arraystretch}{1.1}
	\setlength{\tabcolsep}{4.2pt}
	\begin{tabular}{@{}lccccc|ccccc@{}}
		\toprule
		\multirow{2}{*}{\textbf{Method}} & \multicolumn{5}{c}{\textbf{IM2GPS3k}} & \multicolumn{5}{c}{\textbf{YFCC4k}} \\
		\cmidrule(lr){2-6}\cmidrule(lr){7-11}
		& 1\,km & 25\,km & 200\,km & 750\,km & 2500\,km & 1\,km & 25\,km & 200\,km & 750\,km & 2500\,km \\
		\midrule
		GeoCLIP$^*$~\cite{GeoCLIP23} & \underline{13.1} & 32.2 & 48.0 & 66.5 & \textbf{82.3} & 10.0 & 20.0 & 33.9 & 57.0 & 76.7 \\
		G3$^*$~\cite{G325} & 12.1 & \underline{33.0} & 47.5 & 65.4 & 81.7 & 14.4 & 25.9 & 39.8 & 59.1 & 77.4 \\
		DualGeo$^*$~\cite{DualGeo} & 12.8 & 32.3 & 43.2 & 58.8 & 76.4 & \underline{26.9} & \underline{34.9} & 43.1 & 58.5 & 74.2 \\
		\midrule
		GeoMetric (ours) & \textbf{14.0} & \textbf{36.8} & \textbf{50.6} & \textbf{67.3} & \textbf{82.3} & \textbf{27.0} & \textbf{37.2} & \textbf{47.2} & \textbf{64.3} & \textbf{79.6} \\
		\bottomrule
	\end{tabular}
\end{table}

\subsection{Retrieval Performance Comparison}
\label{sec:retr}

Table~\ref{tab:ret} removes the LMM entirely: each row is the top-1 retrieval of
its own gallery, rebuilt under identical conditions. GeoMetric leads in nine of
the ten cells and ties the tenth, so the gain is already present before any
reasoning takes place.

On YFCC4k, whose scenes are visually ambiguous, the margin grows from street
level to country level and narrows again at continental scale, which is exactly
the range that the anchor prior and the multi-band code target. At 1\,km the
strongest galleries are within a tenth of a point of each other, so this is
where a pure retrieval pipeline saturates and the LMM stage adds its few
tenths.

\begin{table}[t]
	\centering
	\caption{\textbf{Component ablation, retrieval only.} Text: text encoder;
		Multi-Scale: the three-band capsule bank; Anchor: GeoAnchorAttention; Mask:
		the pair of false-negative masks. Accuracies (\%) are at a geodesic error
		threshold of 1, 25, 200, 750 and 2{,}500\,km; Med.\ is the median error in
		km. Best per column in bold, second best underlined.}
	\label{tab:abl}
	\scriptsize
	\vspace{2pt}
	\renewcommand{\arraystretch}{1.1}
	\setlength{\tabcolsep}{2.2pt}
	\begin{tabular}{@{}lcccccccccccccccc@{}}
		\toprule
		\multirow{2}{*}{\textbf{Variant}} & \multirow{2}{*}{\textbf{Text}} & \multirow{2}{*}{\shortstack{\textbf{Multi-}\\\textbf{Scale}}} & \multirow{2}{*}{\textbf{Anchor}} & \multirow{2}{*}{\textbf{Mask}} & \multicolumn{6}{c}{\textbf{IM2GPS3k}} & \multicolumn{6}{c}{\textbf{YFCC4k}} \\
		\cmidrule(lr){6-11}\cmidrule(lr){12-17}
		& & & & & 1 & 25 & 200 & 750 & 2500 & Med. & 1 & 25 & 200 & 750 & 2500 & Med. \\
		\midrule
		w/o Text        & \cx & \ck & \ck & \ck & 12.8 & 34.7 & 48.0 & 65.7 & 81.9 & 235.2 & 26.2 & 36.6 & 45.9 & 63.3 & 78.5 & 280.9 \\
		w/o Multi-Scale & \ck & \cx & \ck & \ck & 13.2 & 36.0 & 49.7 & \textbf{67.3} & \underline{82.1} & 204.3 & 25.1 & 34.9 & 44.9 & 63.1 & \underline{79.8} & 305.8 \\
		w/o Anchor      & \ck & \ck & \cx & \ck & 13.0 & 35.8 & 49.9 & 66.7 & 81.8 & 200.1 & 26.1 & 35.8 & 45.6 & 63.6 & 79.3 & 285.7 \\
		w/o Mask        & \ck & \ck & \ck & \cx & \underline{13.6} & \textbf{36.9} & \underline{50.3} & 67.0 & \underline{82.1} & \underline{191.3} & \underline{26.8} & \underline{36.9} & \underline{46.8} & \textbf{64.5} & \textbf{80.0} & \underline{266.4} \\
		\midrule
		Full & \ck & \ck & \ck & \ck & \textbf{14.0} & \underline{36.8} & \textbf{50.6} & \textbf{67.3} & \textbf{82.3} & \textbf{185.5} & \textbf{27.0} & \textbf{37.2} & \textbf{47.2} & \underline{64.3} & 79.6 & \textbf{263.5} \\
		\bottomrule
	\end{tabular}
\end{table}

\subsection{Component Ablation} 
\label{sec:abl}

Table~\ref{tab:abl} removes one component at a time, with the LMM disabled so that only the retrieval representation is measured. No variant matches the full model (w/o LMM) on both threshold accuracy and median error, and the four ablations degrade different parts of the error curve.

\textit{The text encoder carries the coarse geographic knowledge.} Removing it drops the 25- and 200-km accuracy on IM2GPS3k by more than two points each and raises the median error by about 50\,km, the largest median increase on this benchmark; its effect is concentrated at the city-to-region scale, where appearance alone is weakly informative.

\textit{The multi-scale bank keeps visually similar scenes apart.} Its removal hurts most on YFCC4k, the benchmark of ambiguous scenes, lowering street- and city-level accuracy and raising the median by more than 40\,km. Without the three-band representations, nearby locations that share local appearance are harder to separate.

\textit{GeoAnchorAttention provides a global prior that complements the local code.} Removing it reduces finest-threshold accuracy on IM2GPS3k and city-to-region accuracy on YFCC4k, consistent with a prior useful where the local code is ambiguous.

\textit{The masks act on the error tail rather than the threshold curve.} Removing them leaves threshold accuracy within half a point of the full model, and at some thresholds it is marginally higher, but the median error rises on both benchmarks. As predicted by the debiasing argument in Sect.~\ref{sec:trimodal}, the masks do not lift the threshold curve; they down-weight false negatives that pull matched pairs apart, shortening the tail.

The four components are complementary and operate at different scales: the text encoder supplies coarse semantics, the multi-scale bank separates fine-grained scenes, the anchor attention supplies a global prior, and the masks control tail errors.

\begin{figure}[t]
	\centering
	\includegraphics[width=\linewidth]{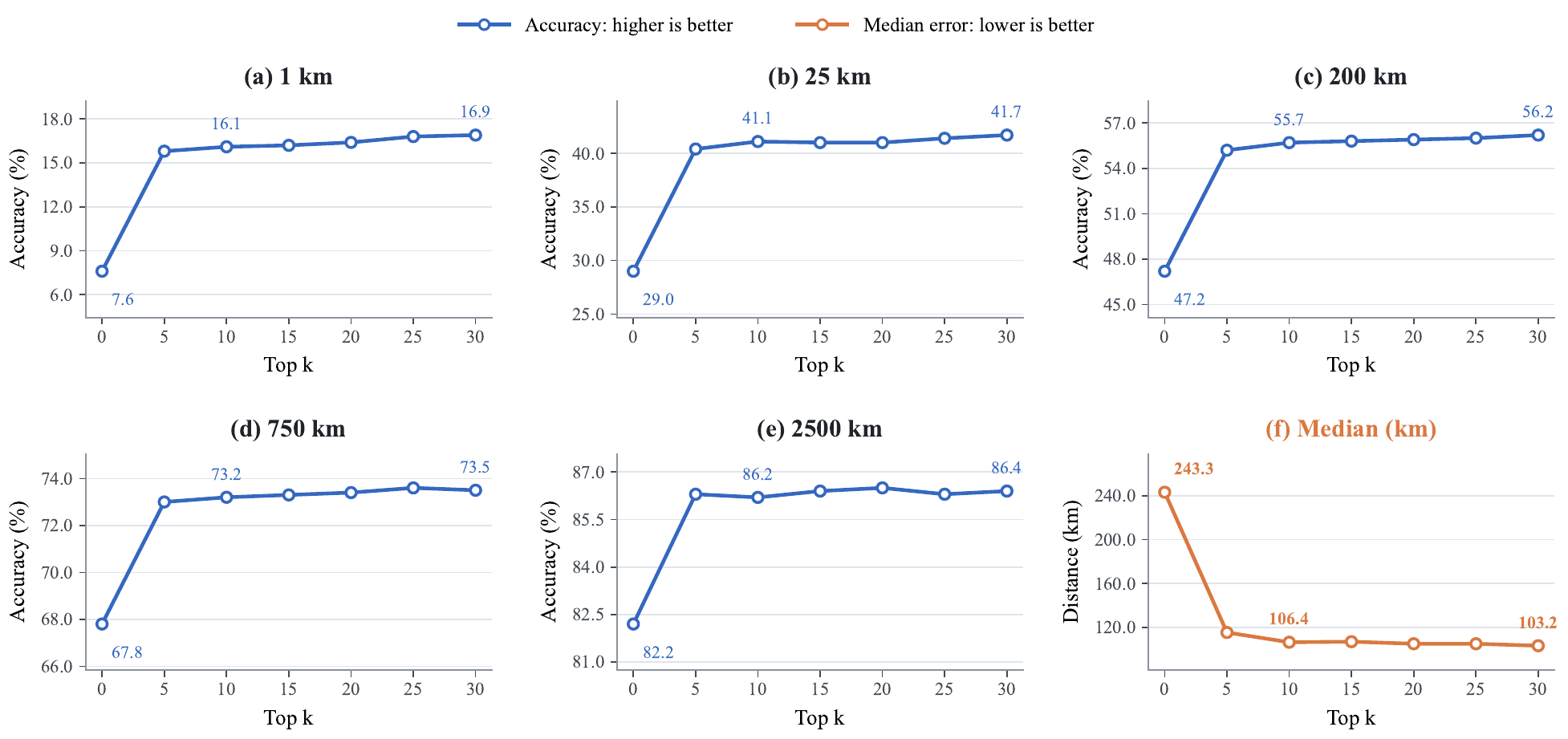}
	\caption{\textbf{Candidate budget on IM2GPS3k.} The geographic text is
		disabled ($g{=}\text{none}$), so only coordinates reach the LMM and $k{=}0$ is
		direct inference without retrieval. Panels (a)--(e): accuracy at the five
		thresholds; (f): median error. Accuracy rises steeply up to $k{=}5$ and
		saturates; beyond $k{=}10$ the gains are within $0.8$ points.}
	\label{fig:cand}
\end{figure}

\subsection{Hyperparameter and LMM Backbone Analysis}
\label{sec:inf}

\paragraph{Candidate budget.} Fig.~\ref{fig:cand} varies $k$ with the geographic
text switched off, so that only coordinates reach the LMM. Without retrieval
($k{=}0$) the model still finds the right continent but almost never the street:
at 1\,km it reaches about half of what retrieval alone achieves, whereas at
750 and 2{,}500\,km the two are on par. Five candidates already close most of the
gap, and beyond $k{=}10$ the curves are flat, with the remaining gain under a
point at three times the prompt length. We therefore fix $k{=}10$.

\begin{table}[h]
	\centering
	\caption{\textbf{Geographic text at $k{=}10$.} Each row lists the fields that are
		present on top of the candidate coordinates, and the last row adds all three. Accuracies (\%) are at a geodesic error
		threshold of 1, 25, 200, 750 and 2{,}500\,km; Med.\ is the median error in
		km. Best per column in \textbf{bold}.}
	\label{tab:gran}
	\scriptsize
	\vspace{2pt}
	\renewcommand{\arraystretch}{1.0}
	\begin{tabular*}{\linewidth}{@{\extracolsep{\fill}}cccccccccc@{}}
		\toprule
		\multicolumn{4}{c}{\textbf{Granularity}} & \multicolumn{6}{c}{\textbf{IM2GPS3k}} \\
		\cmidrule(lr){1-4}\cmidrule(lr){5-10}
		\textbf{Continent} & \textbf{Country} & \textbf{City} & \textbf{GPS} & 1\,km & 25\,km & 200\,km & 750\,km & 2500\,km & Med. \\
		\midrule
		\cx & \cx & \cx & \ck & 16.1 & 41.1 & 55.7 & 73.2 & 86.2 & 106.4 \\
		\ck & \cx & \cx & \ck & 16.3 & 41.1 & 55.6 & 73.2 & 86.1 & 110.1 \\
		\cx & \ck & \cx & \ck & 16.4 & 41.2 & 56.0 & 73.4 & 86.3 & 106.8 \\
		\cx & \cx & \ck & \ck & \textbf{17.1} & \textbf{41.7} & \textbf{56.2} & \textbf{73.7} & \textbf{86.4} & \textbf{102.9} \\
		\ck & \ck & \ck & \ck & 16.8 & 41.4 & 55.9 & 73.3 & 85.8 & 106.0 \\
		\bottomrule
	\end{tabular*}
\end{table}

\paragraph{Geographic text.} Table~\ref{tab:gran} fixes $k{=}10$ and adds one
field at a time. Adding the city raises every threshold and gives the lowest
median; the country and the continent are close to neutral, and the full
three-field text is actually worse than the city alone. The coordinate already
fixes the coarse region, so country and continent tokens add little, while the
city name is the field that carries information the coordinate does not.

\paragraph{Batch size.} Table~\ref{tab:ab}a compares an effective batch of 1024
with 2048. Every threshold is within $0.7$ points and the ordering between the
two sizes is not monotone, which is the behavior Sect.~\ref{sec:geoanchor}
predicts: the location encoder never mixes samples, so the batch only changes the
number of negatives and not the geometry of the code.

\paragraph{LMM backbone.} Table~\ref{tab:ab}b repeats the evaluation with
Qwen2.5-VL-7B~\cite{qwen25vl}, holding the retrieval pool, $k$ and the prompt
fixed. The two backbones agree within $1.1$ points at every threshold and
neither dominates: the 7B model is ahead at 25 and 200\,km ($42.3$ and $56.8$
against $41.7$ and $56.2$), the 8B model at 1, 750 and 2{,}500\,km. The
advantage of the candidate context therefore does not depend on the particular
LMM, and a lightweight open model is already sufficient.

\begin{table}[t]
	\centering
	\caption{\textbf{(a) Effective batch size} and \textbf{(b) LMM backbone} on
	IM2GPS3k; the retrieval pool, $k$ and the prompt are identical. Best in bold.}
	\label{tab:ab}
	\scriptsize
	\vspace{2pt}
	\begin{subtable}[t]{0.44\linewidth}
		\centering
		\renewcommand{\arraystretch}{1.0}
		\setlength{\tabcolsep}{3pt}
		\begin{tabular}{@{}lccccc@{}}
			\toprule
			\textbf{Batch} & 1\,km & 25\,km & 200\,km & 750\,km & 2500\,km \\
			\midrule
			eff-1024 & 13.9 & 36.7 & 49.9 & \textbf{67.4} & \textbf{82.4} \\
			eff-2048 & \textbf{14.0} & \textbf{36.8} & \textbf{50.6} & 67.3 & 82.3 \\
			\bottomrule
		\end{tabular}
		\caption{Effective batch size.}
	\end{subtable}\hfill
	\begin{subtable}[t]{0.5\linewidth}
		\centering
		\renewcommand{\arraystretch}{1.0}
		\setlength{\tabcolsep}{3pt}
		\begin{tabular}{@{}lccccc@{}}
			\toprule
			\textbf{Backbone} & 1\,km & 25\,km & 200\,km & 750\,km & 2500\,km \\
			\midrule
			Qwen3-VL-8B & \textbf{17.1} & 41.7 & 56.2 & \textbf{73.7} & \textbf{86.4} \\
			Qwen2.5-VL-7B & 16.1 & \textbf{42.3} & \textbf{56.8} & 73.2 & 85.3 \\
			\bottomrule
		\end{tabular}
		\caption{LMM backbone.}
	\end{subtable}
\end{table}

\subsection{Qualitative Comparison}
\label{sec:vis}

\begin{figure}[t]
	\centering
	\includegraphics[width=0.95\linewidth]{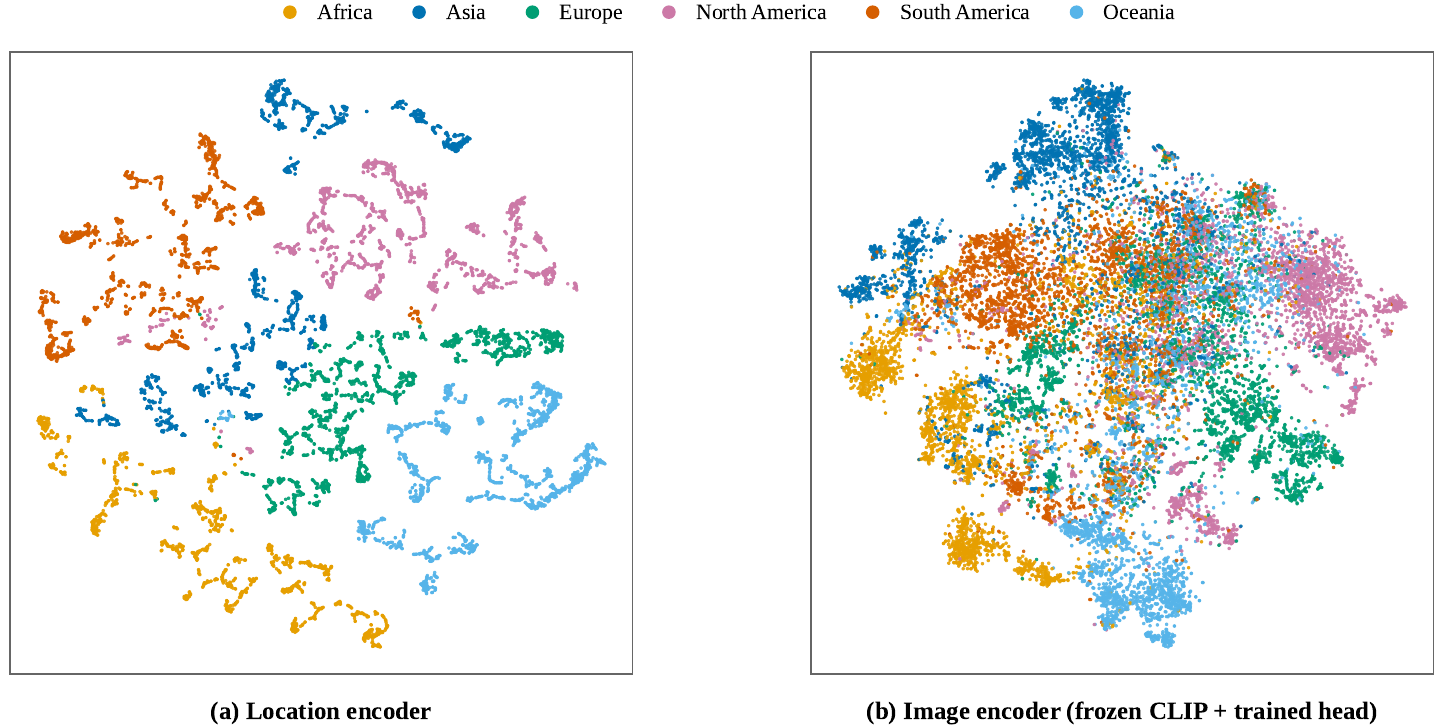}
	\caption{\textbf{Geographic structure in the two modalities.} t-SNE of
		18{,}000 MP16-Pro~\cite{G325} samples (3{,}000 per continent), colored by
		continent. (a) Coordinates from the trained location encoder: continents form
		compact, largely separated groups. (b) CLIP ViT-L/14-336 features through the
		trained image encoder: continent structure is only partial, and the continents
		overlap substantially. The two panels are separate t-SNE embeddings, so axes and inter-cluster distances are not comparable; the comparison is qualitative: whether continents separate within each panel.}
	\label{fig:tsne}
\end{figure}

Fig.~\ref{fig:tsne} visualizes the two modalities in a two-dimensional projection
of the representation used for retrieval. We sample 18{,}000 MP16-Pro~\cite{G325}
images, 3{,}000 per continent, and project (a) coordinates encoded by the trained
location encoder and (b) CLIP features passed through the trained image encoder,
coloring both by continent.

The two panels show different organizations. In (a), samples from the same
continent form compact groups and the continents are largely separated, so the
location encoder exposes continent-level geographic structure in the shared
space. In (b), the continents overlap substantially: some peripheral groups are
visible, but no clean continent-level partition emerges. The image encoder thus
retains only partial geographic structure in this projection. This contrast is
consistent with the design, where the location encoder supplies the explicit
geographic frame and training aligns the image encoder to it. The comparison is
qualitative and does not show that the image encoder lacks geographic information;
the projection exposes only part of each representation.

\section{Conclusion}
\label{sec:conclusion}

We presented GeoMetric, a retrieval framework that makes geographic distance part
of the representation rather than something applied to labels, prompts or
reranking scores. In Phase~I, image, text and location embeddings are learned in
a unit hypersphere. ScaleFusion projects a coordinate onto the Equal Earth chart
and lifts it with three Fourier bands, fusing the bands by attention into a
multi-scale embedding that spans street to continental scale. GeoAnchorAttention adds a global prior from a fixed set of spherical anchors
without attending across samples, so each coordinate embedding is
batch-independent and the gallery can be indexed once. A trimodal contrastive
objective aligns images, reverse-geocoded texts and coordinates, with semantic
and geographic masks that remove false negatives. Phase~II retrieves candidates, each carrying a GPS coordinate and
granularity-aware geographic text; Phase~III lets a frozen open-source 8B LMM
choose the coordinate. Across four benchmarks, GeoMetric is best in 15 of 20
threshold cells with an open 8B LMM, and remains competitive at every geographic
scale. Ablations show that the components act at different scales: text carries
coarse geographic context, the multi-scale bank separates similar scenes, the
anchor supplies a global frame, and the masks shape the error distribution. These
results indicate that geographic structure is most effective when learned in the
coordinate encoder, where it shapes both retrieval and the evidence given to the
LMM.

\textit{The code will be released upon acceptance.}

\bibliographystyle{splncs04}
\bibliography{GeoMetric}

@inproceedings{Vaswani17,
	author    = {Vaswani, A. and Shazeer, N. and Parmar, N. and Uszkoreit, J. and Jones, L. and Gomez, A. and Kaiser, L. and Polosukhin, I.},
	title     = {Attention is all you need},
	booktitle = {Advances in Neural Information Processing Systems (NeurIPS)},
	volume    = {30},
	pages     = {5998--6008},
	year      = {2017},
	doi       = {10.48550/arXiv.1706.03762}
}

@inproceedings{Rahimi07,
	author    = {Rahimi, A. and Recht, B.},
	title     = {Random features for large-scale kernel machines},
	booktitle = {Advances in Neural Information Processing Systems (NeurIPS)},
	volume    = {20},
	pages     = {1177--1184},
	year      = {2007}
}

@article{Johnson19,
	author  = {Johnson, J. and Douze, M. and J{\'e}gou, H.},
	title   = {Billion-scale similarity search with {GPUs}},
	journal = {IEEE Transactions on Big Data},
	volume  = {7},
	number  = {3},
	pages   = {535--547},
	year    = {2021},
	doi     = {10.1109/TBDATA.2019.2921572}
}

@article{Choi22,
	author  = {Choi, J. and Wong, C. and Hajj-Ahmad, A. and Wu, M. and Ren, Y.},
	title   = {Invisible Geolocation Signature Extraction From a Single Image},
	journal = {IEEE Transactions on Information Forensics and Security},
	volume  = {17},
	pages   = {2598--2613},
	year    = {2022},
	doi     = {10.1109/TIFS.2022.3185775}
}

@inproceedings{Clip21,
	title={{Learning transferable visual models from natural language supervision}},
	author={Radford, A. and Kim, J. W. and Hallacy, C. and Ramesh, A. and Goh, G. and Agarwal, S. and Sastry, G. and Askell, A. and Mishkin, P. and Clark, J. and Krueger, G. and Sutskever, I.}, 
	booktitle={International Conference on Machine Learning (ICML)},
	pages={8748--8763},
	year={2021},
	doi       = {10.48550/arXiv.2103.00020}
}

@inproceedings{CPlaNet18,
	author    = {P. H. Seo and T. Weyand and J. Sim and B. Han},
	title     = {{CPlaNet: Enhancing image geolocalization by combinatorial partitioning of maps}},
	booktitle = {European Conference on Computer Vision (ECCV)},
	pages     = {544--560},
	year      = {2018},
	doi       = {10.1007/978-3-030-01249-6_33}
}

@article{Cui27,
	author  = {Cui, J. and Luo, X. and Shi, W. and Ma, X. and Wu, N. and Li, H.},
	title   = {Tracing Locations from Pixels: A Survey of Worldwide Image Geo-Localization},
	journal = {Information Processing and Management},
	volume  = {64},
	pages   = {105226},
	year    = {2027},
	doi     = {10.1016/j.ipm.2026.105226}
}

@inproceedings{GeoSURGE,
	author = {A. Daruna and N. Meegan and H. Chiu and S. Samarasekera and R. Kumar},
	title = {{GeoSURGE: Geo-localization using Semantic Fusion with Hierarchy of Geographic Embeddings}},
	year = {2026},
	pages = {41396--41405},
	booktitle = {Proceedings of the IEEE/CVF Conference on Computer Vision and Pattern Recognition (CVPR)},
	doi       = {10.48550/arXiv.2510.01448}
}

@inproceedings{HierLoc26,
	author    = {Gadi, H.K. and Matos, D. and Luo, H. and Liu, L. and Wang, Y. and Zhang, Y. and Meng, L.},
	title     = {{HierLoc}: Hyperbolic Entity Embeddings for Hierarchical Visual Geolocation},
	booktitle = {Proceedings of the International Conference on Learning Representations (ICLR)},
	year      = {2026},
	doi       = {10.48550/arXiv.2601.23064}
}

@inproceedings{Haas24,
	author = {L. Haas and M. Skreta and S. Alberti and C. Finn},
	title     = {{PIGEON}: Predicting Image Geolocations},
	booktitle = {Proceedings of the IEEE/CVF Conference on Computer Vision and Pattern Recognition (CVPR)},
	pages     = {12893--12902},
	year      = {2024},
	doi       = {10.1109/CVPR52733.2024.01225}
}

@inproceedings{IM2GPS08,
	author    = {Hays, J. and Efros, A.},
	title     = {{IM2GPS}: Estimating Geographic Information from a Single Image},
	booktitle = {Proceedings of the IEEE/CVF Conference on Computer Vision and Pattern Recognition (CVPR)},
	pages     = {1--8},
	year      = {2008},
	doi       = {10.1109/CVPR.2008.4587784}
}

@inproceedings{SpotAgent,
	author = {F. Jia and L.  Dai and W. Deng and F. Zhang and C. Hu and D. Jiang and Y. Liu},
	title     = {{SpotAgent}: Grounding Visual Geo-Localization in Large Vision-Language Models Through Agentic Reasoning},
	booktitle = {Proceedings of the ACM SIGKDD Conference on Knowledge Discovery and Data Mining (KDD)},
	pages     = {2062--2073},
	year      = {2026},
	doi       = {10.1145/3770855.3817948}
}

@inproceedings{G325,
	author    = {Jia, P. and Liu, Y. and Li, X. and Wang, Y. and Du, Y. and Han, X. and Wei, X. and Wang, S. and Yin, D. and Zhao, X.},
	title     = {{G3}: An Effective and Adaptive Framework for Worldwide Geolocalization Using Large Multi-Modality Models},
	booktitle = {Advances in Neural Information Processing Systems (NeurIPS)},
	volume    = {37},
	pages     = {53198--53221},
	year      = {2024},
	doi       = {10.52202/079017-1685}
}

@inproceedings{GeoRanker25,
	author    = {Jia, P. and Park, S. and Gao, S. and Zhao, X. and Li, S.},
	title     = {{GeoRanker}: Distance-Aware Ranking for Worldwide Image Geolocalization},
	booktitle = {Advances in Neural Information Processing Systems (NeurIPS)},
	volume    = {38},
	pages     = {20100--20126},
	year      = {2025},
	doi       = {10.52202/085713-0596}
}

@inproceedings{GeoCLIP23,
	author    = {Cepeda, V. and Nayak, G. and Shah, M.},
	title     = {{GeoCLIP}: {CLIP}-Inspired Alignment Between Locations and Images for Effective Worldwide Geo-Localization},
	booktitle = {Advances in Neural Information Processing Systems (NeurIPS)},
	volume    = {36},
	pages     = {8690--8701},
	year      = {2023},
	doi       = {10.52202/075280-0379}
}

@inproceedings{GeoReasoner,
	author    = {Li, L. and Ye, Y. and Jiang, B. and Zeng, W.},
	title     = {{GeoReasoner: Geo-Localization with Reasoning in Street Views Using a Large Vision-Language Model}},
	booktitle = {Proceedings of the International Conference on Machine Learning (ICML)},
	pages     = {29222--29233},
	year      = {2024},
	doi       = {10.48550/arXiv.2406.18572}
}

@inproceedings{ISNs18,
	author    = {M{\"u}ller-Budack, E. and Pustu-Iren, K. and Ewerth, R.},
	title     = {Geolocation Estimation of Photos Using a Hierarchical Model and Scene Classification},
	booktitle = {Proceedings of the European Conference on Computer Vision (ECCV)},
	pages     = {575--592},
	year      = {2018},
	doi       = {10.1007/978-3-030-01258-8_35}
}

@inproceedings{TransLocator22,
	author    = {Pramanick, S. and Nowara, E. and Gleason, J. and Castillo, C. and Chellappa, R.},
	title     = {Where in the World Is This Image? Transformer-Based Geo-Localization in the Wild},
	booktitle = {Proceedings of the European Conference on Computer Vision (ECCV)},
	pages     = {196--215},
	year      = {2022},
	doi       = {10.1007/978-3-031-19839-7_12}
}

@inproceedings{GeoBayes,
	author    = {Shi, W. and Li, X. and Li, K. and Fang, J. and Zhou, Q. and Geng, Q. and Zhou, Z.},
	title     = {{GeoBayes}: Probabilistic Image Geo-Localization Inference via Sequential Bayesian Updating},
	booktitle = {Proceedings of the AAAI Conference on Artificial Intelligence (AAAI)},
	pages     = {8997--9005},
	year      = {2026},
	doi       = {10.1609/aaai.v40i11.37855}
}

@inproceedings{Semp22,
	author    = {Theiner, J. and M{\"u}ller-Budack, E. and Ewerth, R.},
	title     = {Interpretable Semantic Photo Geolocation},
	booktitle = {Proceedings of the IEEE/CVF Winter Conference on Applications of Computer Vision (WACV)},
	pages     = {1474--1484},
	year      = {2022},
	doi       = {10.1109/WACV51458.2022.00154}
}

@inproceedings{Vo17,
	author    = {Vo, N. and Jacobs, N. and Hays, J.},
	title     = {Revisiting {IM2GPS} in the Deep Learning Era},
	booktitle = {Proceedings of the IEEE/CVF International Conference on Computer Vision (ICCV)},
	pages     = {2640--2649},
	year      = {2017},
	doi       = {10.1109/ICCV.2017.286}
}

@inproceedings{GRE,
	author    = {Wang, C. and Ye, X. and Pan, X. and Pan, Z. and Wang, H. and Song, Y.},
	title     = {{GRE Suite}: Geo-Localization Inference via Fine-Tuned Vision-Language Models and Enhanced Reasoning Chains},
	booktitle = {Advances in Neural Information Processing Systems (NeurIPS)},
	volume    = {38},
	pages     = {102548--102578},
	year      = {2025},
	doi       = {10.52202/085713-3091}
}

@inproceedings{PlaNet16,
	author    = {Weyand, T. and Kostrikov, I. and Philbin, J.},
	title     = {{PlaNet} -- Photo Geolocation with Convolutional Neural Networks},
	booktitle = {Proceedings of the European Conference on Computer Vision (ECCV)},
	pages     = {37--55},
	year      = {2016},
	doi       = {10.1007/978-3-319-46484-8_3}
}

@inproceedings{DASFAA25ANN,
	author    = {Zhang, Z. and Shang, J. and Li, K. and Li, J. and Tian, M. and Wang, X.},
	title     = {Learning Distance-Aware Space Partitions for Approximate Nearest Neighbor Search},
	booktitle = {Database Systems for Advanced Applications (DASFAA)},
	series    = {Lecture Notes in Computer Science},
	volume    = {15989},
	pages     = {140--156},
	publisher = {Springer},
	year      = {2025},
	doi       = {10.1007/978-981-95-4149-2_9}
}

@inproceedings{DASFAA25LASE,
	author    = {Cai, P. and Ma, C. and Lu, Z. and Liu, P. and Liu, P. and Li, C. and Chen, H.},
	title     = {{LASE}: A Learned Spatial Index for Dynamic Workloads},
	booktitle = {Database Systems for Advanced Applications (DASFAA)},
	series    = {Lecture Notes in Computer Science},
	volume    = {15989},
	pages     = {72--87},
	publisher = {Springer},
	year      = {2025},
	doi       = {10.1007/978-981-95-4149-2_5}
}

@inproceedings{DASFAA25GNN,
	author    = {An, Y. and Tong, S. and Wang, Q. and Zhu, R. and Zhang, A. and Zong, C. and Jiang, H. and Wang, B.},
	title     = {Dynamic Group Nearest Neighbor Group Query over Streaming Data},
	booktitle = {Database Systems for Advanced Applications (DASFAA)},
	series    = {Lecture Notes in Computer Science},
	volume    = {15989},
	pages     = {421--437},
	publisher = {Springer},
	year      = {2025},
	doi       = {10.1007/978-981-95-4149-2_28}
}

@inproceedings{Img2loc24,
	author    = {Zhou, Z. and Zhang, J. and Guan, Z. and Hu, M. and Lao, N. and Mu, L. and Li, S. and Mai, G.},
	title     = {{Img2Loc}: Revisiting Image Geolocalization Using Multi-Modality Foundation Models and Image-Based Retrieval-Augmented Generation},
	booktitle = {Proceedings of the International ACM SIGIR Conference on Research and Development in Information Retrieval (SIGIR)},
	pages     = {2749--2754},
	year      = {2024},
	doi       = {10.1145/3626772.3657673}
}

@inproceedings{GeoDecoder23,
	author    = {Clark, B. and Kerrigan, A. and Kulkarni, P. and Cepeda, V. and Shah, M.},
	title     = {Where We Are and What We're Looking At: Query Based Worldwide Image Geo-Localization Using Hierarchies and Scenes},
	booktitle = {Proceedings of the IEEE/CVF Conference on Computer Vision and Pattern Recognition (CVPR)},
	pages     = {23182--23190},
	year      = {2023},
	doi       = {10.1109/CVPR52729.2023.02220}
}

@article{Savric19,
	author  = {{\v S}avri{\v c}, Bojan and Patterson, Tom and Jenny, Bernhard},
	title   = {The {Equal Earth} Map Projection},
	journal = {International Journal of Geographical Information Science},
	volume  = {33},
	number  = {3},
	pages   = {454--465},
	year    = {2019},
	doi     = {10.1080/13658816.2018.1504949}
}

@article{Gonzalez10,
	author  = {Gonz{\'a}lez, {\'A}lvaro},
	title   = {Measurement of Areas on a Sphere Using {Fibonacci} and Latitude--Longitude Lattices},
	journal = {Mathematical Geosciences},
	volume  = {42},
	number  = {1},
	pages   = {49--64},
	year    = {2010},
	doi     = {10.1007/s11004-009-9257-x}
}

@inproceedings{Chuang20,
	author    = {Chuang, C. and Robinson, J. and Lin, Y.-C. and Torralba, A. and Jegelka, S.},
	title     = {Debiased Contrastive Learning},
	booktitle = {Advances in Neural Information Processing Systems (NeurIPS)},
	volume    = {33},
	pages     = {8765--8775},
	year      = {2020},
	doi       = {10.48550/arXiv.2007.00224}
}

@misc{Qwen3VL,
	author = {{Qwen Team}},
	title  = {{Qwen3-VL} Technical Report},
	year   = {2025},
	doi    = {10.48550/arXiv.2511.21631},
	note   = {Alibaba Group}
}

@article{qwen25vl,
	author  = {{Qwen Team}},
	title   = {{Qwen2.5-VL} Technical Report},
	journal = {arXiv preprint arXiv:2502.13923},
	year    = {2025},
	doi     = {10.48550/arXiv.2502.13923}
}

@misc{DualGeo,
	author       = {Cui, J. and Shi, W. and Du, S. and He, H. and Ma, X. and Tang, H. and Luo, X.},
	title        = {{DualGeo}: A Dual-View Framework for Worldwide Image Geo-Localization},
	year         = {2026},
	howpublished = {arXiv:2604.25533},
	doi          = {10.48550/arXiv.2604.25533},
	note         = {Accepted at ICME 2026}
}

@inproceedings{CAMeRA,
	author    = {Hu, B. and Li, Y. and Cheng, B.},
	title     = {{CAMeRA}: A Mamba-Based Context-Aware Adaptive Multimodal Architecture for Sequential Recommendation},
	booktitle = {Database Systems for Advanced Applications (DASFAA)},
	series    = {Lecture Notes in Computer Science},
	volume    = {16535},
	pages     = {171--187},
	publisher = {Springer},
	year      = {2026},
	doi       = {10.1007/978-981-92-0363-5_11}
}

@inproceedings{GeoPhraseTree,
	author    = {Wu, C. and Lu, C. and Jing, Y. and He, Z. and Yang, Z. and Wang, X.S.},
	title     = {{GeoPhrase Tree}: An Efficient Index for Frequent Phrase Query over Spatio-Temporal Ranges},
	booktitle = {Database Systems for Advanced Applications (DASFAA)},
	series    = {Lecture Notes in Computer Science},
	volume    = {16535},
	pages     = {3--19},
	publisher = {Springer},
	year      = {2026},
	doi       = {10.1007/978-981-92-0363-5_1}
}

\end{document}